# Investigating the Effect of Spatial Context on Multi-Task Sea Ice Segmentation

Behzad Vahedi [1], Rafael Pires de Lima[1], Sepideh Jalayer [1], Walter N. Meier [2], Andrew P. Barrett [2], Morteza Karimzadeh [1]

1 Department of Geography, University of Colorado Boulder, Boulder, CO, USA

2 National Snow and Ice Data Center (NSIDC), CIRES, University of Colorado Boulder, Boulder, CO, USA

**Abstract**

Capturing spatial context at multiple scales is crucial for deep learning-based sea ice segmentation. However, the optimal specification of spatial context based on observation resolution and task characteristics remains underexplored. This study investigates the impact of spatial context on the segmentation of sea ice concentration, stage of development, and floe size using a multi-task segmentation model. We implement Atrous Spatial Pyramid Pooling with varying atrous rates to systematically control the receptive field size of convolutional operations, and to capture multi-scale contextual information. We explore the interactions between spatial context and feature resolution for different sea ice properties and examine how spatial context influences segmentation performance across different input feature combinations from Sentinel-1 SAR and Advanced Microwave Radiometer-2 (AMSR2) for multi-task mapping. Using Gradient-weighted Class Activation Mapping, we visualize how atrous rates influence model decisions. Our findings indicate that smaller receptive fields excel for high-resolution Sentinel-1 data, while medium receptive fields yield better performances for stage of development segmentation and larger receptive fields often lead to diminished performances. The fusion of SAR and AMSR2 enhances segmentation across all tasks. We highlight the value of lower-resolution 18.7 and 36.5 GHz AMSR2 channels in sea ice mapping. These findings highlight the importance of selecting appropriate spatial context based on observation resolution and target properties in sea ice mapping. By systematically analyzing receptive field effects in a multi-task setting, our study provides insights for optimizing deep learning models in geospatial applications.

1. **Introduction**

Capturing multi-scale contextual information is beneficial for modeling complex environmental phenomena such as land cover, vegetation, or urban infrastructure, where interactions between fine-scale details and broader spatial patterns influences segmentation, classification, and object detection tasks (Chen et al., 2020; Li et al., 2014; Yang et al., 2023; Zhang et al., 2020). By capturing spatial context and dependencies across multiple scales, models can achieve improved performances particularly when integrating data sources with different resolutions. In sea ice mapping, a multi-scale or multi-context approach is essential due to the complex spatial structure of sea ice and its properties such as stage of development or concentration (Song et al., 2022). The importance of automated sea ice mapping has grown significantly, driven by the critical role of sea ice in climate change, environmental monitoring, and marine navigation. Traditional manual sea ice mapping methods are labor-intensive and time-consuming, and thus, cannot satisfy the need for high-resolution timely sea ice maps. This has led to a recent surge in research on deep learning-based sea ice classification methods to map different sea ice characteristics such as Sea Ice Concentration (SIC) (Cooke and Scott, 2019; de Gélis et al., 2021; Radhakrishnan et al., 2021; Stokholm et al., 2022), Stage of Development (SOD) (Huang et al., 2024; Pires de Lima et al., 2023; Vahedi et al., 2024), floe size (Nagi et al., 2021; Wang et al., 2024), and thickness (Y. Zhao et al., 2023).

Synthetic Aperture Radar (SAR) imagery has been the primary input for deep-learning-based sea ice mapping and classification research. SAR sensors, such as the Sentinel-1 platform, offer high spatial resolution and all-weather imaging capabilities, making them ideal for capturing detailed surface features in polar regions. However, SAR imagery is often affected by wind conditions and thermal noise. Moreover, different ice types can have similar signatures in SAR backscatter, making it difficult to distinguish them. Therefore, researchers have increasingly focused on fusing SAR imagery with passive microwave radiometer data, particularly from the JAXA Advanced Microwave Scanning Radiometer-2 (AMSR2) sensor (X. Chen, et al., 2024b; Khachatrian et al., 2023; L. Zhao et al., 2023).

Unlike Sentinel-1 SAR, which has a 12-day revisit cycle for a single satellite and a 6-day cycle when both Sentinel-1A and Sentinel-1B were operational (before Sentinel-1B became inoperative on December 2021), AMSR2 has an average revisit frequency of one day. This high temporal resolution enables monitoring of highly changeable sea ice cover and timely updates on changes in ice extent and properties; therefore, it is crucial for both operational forecasting and long-term climate studies (Chi and Kim, 2021; Ivanova et al., 2015).

Such fusion approaches aim to leverage the high-resolution and surface backscatter of SAR with the extensive coverage and temporal dynamics captured by MWR, to enhance the accuracy and robustness of deep learning models for sea ice classification. The AutoICE Challenge Dataset (Stokholm et al., 2024), which contains co-located Sentinel-1 SAR dual-polarized images in addition to AMSR2 brightness temperatures as well as numerical weather prediction parameters from ECMWF Reanalysis v5 has provided a valuable multi-source benchmark of data for training and evaluating deep learning models.

Many studies on sea ice mapping and classification employ Convolutional Neural Network (CNN)-based architectures, such as the ResNet (He et al., 2016) or U-Net (Ronneberger et al., 2015) family of architectures, leveraging their ability to learn local feature representations from input imagery (Khachatrian et al., 2023; Malmgren-Hansen et al., 2021; Radhakrishnan et al., 2021; Wan et al., 2023; L. Zhao et al., 2023). Despite the promise of CNNs, automated sea ice mapping using these architectures faces challenges related to context and scale. The limited field of view inherent in CNNs can restrict their ability to capture the broader spatial context necessary for accurate classification (Dai et al., 2017). Additionally, there is a significant difference in the resolution of Sentinel-1 (93×87 meters with a pixel spacing of 40 meters) and AMSR2 (ranging from 3 to 35 kilometers) images, complicating the fusion of these features and requiring further investigation into the optimal context needed for each image type to achieve the best performance.

Wulf et al. (2024) employed a modified U-Net architecture to classify sea ice concentration by fusing Sentinel-1 SAR and AMSR2 data. Similar to our study, the authors focused on the generalization of their algorithm and uncertainty quantification rather than architectural optimization, therefore, they used a "fairly simple CNN architecture" but were able to achieve pan-Arctic SIC estimation with high accuracy ($R^2$ score of 0.95 on held-out test data). Their results also highlight that the SAR and AMSR2 fusion approach yields consistently higher ice concentration estimates in the marginal ice zone compared to traditional passive microwave products (Wulf et al., 2024).

Stokholm et al. (2022) demonstrated that incorporating a broader context can improve the performance of CNN models in mapping sea ice concentration. They showed that by increasing the receptive field of the U-Net architecture and using larger patches for training, the model can utilize larger-scale spatial information and achieve an improved performance in SIC classification. To increase the receptive field of CNNs, they added additional blocks of convolutional, pooling, and upsampling layers in the encoder and decoder parts of the U-Net, thus making the architecture deeper. This increase in depth, however, also increases the model's complexity, resulting in a higher number of trainable parameters and extended training time, and without reliance on pre-trained methods, increases the need for more training samples to mitigate overfitting. Additionally, they found that using a larger patch size (768×768 pixels) enhanced their model performance. However, they only used Sentinel-1 SAR images as input features in their model and only focused on concentration (SIC).

Song et al. (2022) proposed a method for segmenting sea ice and water in SAR imagery by incorporating multi-scale and contextual information in a model developed based on the Pyramid Scene Parsing Network (PSPNet) architecture (Zhao et al., 2017). Their model includes a multi-scale attention mechanism that fuses features at various scales, capturing both global and local information. However, their model only uses dual-polarized SAR images and performs binary ice-water classification.

Pires de Lima et al. (2023) presented a method to improve sea ice segmentation by integrating atrous convolutions within a customized CNN architecture. Using ResNet for feature extraction and incorporating Atrous Spatial Pyramid Pooling (ASPP) (Chen et al., 2018) in the decoder, their model effectively captures multi-scale contextual information. They showed that this approach enhances sea ice segmentation in both binary ice-water classification and multi-class SOD classification. While this study underscores the importance of scale and context in CNN-based sea ice segmentation applications, only Sentinel-1 SAR features are used as its input features, and it is only applied to sea ice type classification.

Y. Chen et al. (2024) introduced a unified multi-task model called MFDA (Multimodal Fusion Domain Adaptive) to tackle cross-scene sea ice mapping on the AutoICE challenge dataset. MFDA combines Self-Supervised Learning (SSL) for multimodal pre-training, a hybrid convolutional–Transformer architecture for data fusion, and an explicit domain adaptation module to bridge regional differences in sea ice imagery. The authors' focus is on maximizing cross-domain and cross-region generalization by using SSL-driven feature learning and an advanced fusion mechanism to achieve state-of-the-art performance in sea ice segmentation, whereas our approach focuses on investigating the impact of spatial context on segmentation performance. Instead of pre-training or domain adaptation, we train our multi-task CNN model in a fully supervised manner and control multi-scale spatial context via atrous convolutions and feature-level SAR–AMSR2 data fusion to enhance segmentation. Nevertheless, we provide detailed comparisons to MFDA (which is pre-trained using additional larger datasets) in cross-scene evaluation settings.

MFGC-Net proposed by (Ma et al., 2025) is a multi-task encoder-decoder network for sea ice segmentation using the AI4Arctic Sea Ice Challenge Dataset. It uses a ResNet-34 backbone with a cross-scale interaction module that uses cross-attention to enable multi-scale feature fusion in the encoder. To capture long-range dependencies, the authors used a context module combines multi-head self-attention with a channel patch module to enable global spatial and channel-wise context modeling. The decoder in MFGC-Net applies scale-aware fusion and spatial attention to fuse features across different levels. Its multi-scale extraction mechanism is conceptually similar to the ASPP module, but it does not employ parallel dilated convolutions, and instead, utilizes attention-based mechanisms to dynamically extract cross-scale dependencies and therefore does not allow for direct control over the amount of context captured by the model.

Another approach to addressing the limitations of CNNs in capturing contextual information is to downsample Sentinel-1 SAR images. For instance, (X. Chen et al., 2024a; X. Chen et al., 2024b) downsample these images by a factor of 10 to incorporate broader contextual information. However, this downsampling approach adds additional pre-preprocessing and may result in the loss of critical local contextual information.

To mitigate the issues mentioned above, in this study, we extend a multi-task pixel-level semantic segmentation model based on the DeepLab V3 architecture (Chen et al., 2017; Pires de Lima et al., 2023) in which we integrate the ASPP module for multi-task learning. We use this model to segment SIC, SOD, and Floe simultaneously using the AutoICE Challenge dataset (Stokholm et al., 2024).

The segmentation terminology in remote sensing can sometimes refer to partitioning an image into homogeneous regions, without necessarily defining pixel-level class assignment. In a machine learning or deep learning semantic segmentation framework, the model assigns a class label to each pixel in an image, and thus preserves object boundaries and provides predictions at a fine scale. In contrast, image classification predicts a single label for an entire image (or an image patch) without explicitly localizing individual objects, and therefore, it leads to generally coarser predictions. This fundamental distinction has driven the majority of sea ice mapping research towards semantic segmentation frameworks and is the reason why we have chosen the same approach (Chen et al., 2018; X. Chen et al., 2024b; Ronneberger et al., 2015).

Our approach employs atrous (dilated) convolutions similar to (Pires de Lima et al., 2023), albeit with varying atrous rates to capture multi-scale contextual information without downsampling, thereby preserving local details. This method enhances the model's ability to capture both broad and fine-grained contextual information simultaneously.

Additionally, we train models using a fusion of Sentinel-1 SAR backscatter and AMSR2 brightness temperature features, as well as features from only Sentinel-1 SAR or only AMSR2. By using different atrous rates for each feature group, we investigate the effects of contextual information on the performance of the model for each target (SIC, SOD, FLOE) and individual classes. This study aims to provide an analysis of how multi-scale contextual information influence model performance across different sea ice characteristics.

Spatial context and scale play crucial roles in deep learning-based sea ice mapping, particularly when mapping different properties of ice simultaneously using several sources of input data with different resolutions. While these terms are sometimes used interchangeably, it is important to distinguish between them. Scale refers to the granularity or resolution at which sea ice features are analyzed, ranging from fine-scale surface roughness features to collections of floes and open water areas. In traditional remote sensing, spatial context refers to the spatial relationships among neighboring pixels within a single spectral band. This context is often captured using spatial filtering, texture analysis, or statistical measures. Our approach complements this definition by capturing spatial context through the receptive fields of CNNs. Instead of explicitly computing spatial relationships using pre-defined filters or statistical measures, CNNs *learn* spatial dependencies across multiple spatial extents and spectral bands, but indeed capture spatial relationships among neighboring pixels within each spectral band as well.

The key contribution of this study is a systematic investigation of how spatial context influences the segmentation of multiple sea ice properties within a deep learning framework (and not merely a comparison of different input features). Specifically, we examine how the receptive field—controlled by the atrous rate—affects the captured spatial dependencies. It is also important to note that these dependencies are captured not only from the input features but also from the deep feature representations that are generated by the model encoder.

Additionally, we provide explainability by utilizing Grad-CAM analysis and illustrate how the amount of captured spatial context influences model predictions across different tasks. To the best of our knowledge, no previous study has systematically examined the impact of spatial context on sea ice segmentation. Existing studies have primarily focused on the effect of input image size or network depth, without explicitly analyzing how variations in receptive field configurations influence segmentation performance. By isolating the role of spatial context, our work provides new insights into optimizing deep learning models for sea ice mapping.

The specific contributions of our work are:

- We investigate the impact of spatial context and scale, determined by the receptive field size of CNNs by varying atrous rates in the ASPP module, on the overall performance of a multi-task model for classifying SIC, SoD, and FLOE simultaneously. We identify significant changes in overall segmentation performance as well as performance on individual targets, providing insights into the optimal configuration for multi-task learning in sea ice mapping.
- We examine the interaction between SAR and AMSR2 features and assess how these features complement each other in enhancing sea ice segmentation performance across different scales (contexts) in a multi-task setting. This analysis explores the added value of each feature type, particularly in relation to varying spatial context and resolutions.

We describe the AutoICE challenge dataset, the study area, and the distribution of different classes in Section 2. Our methodology, including the model architecture, experimental setup, and training details are provided in Section 3. Our experiments' findings are presented in Section 4, and finally the conclusion of our work is presented in Section 5.

## 2. Data

### 2.1. AutoICE Challenge Dataset

Access to sufficient labeled data has been and continues to be a significant challenge in training reliable deep- learning based sea ice mapping systems. The AutoICE Challenge Dataset (Stokholm et al., 2024), however, has provided a valuable solution for part of the Arctic. This dataset spans a period from January 2018 to December 2021 and includes 533 co-located and georeferenced scenes from the Canadian and Greenlandic Arctic, providing great coverage that is essential for robust model training and validation. Of the 533 scenes, 20 comprise the test set, and the remaining 513 scenes comprise the training set.

Each scene in the dataset is composed of multiple data sources, including: (1) dual-polarized (HH and HV) Sentinel-1 C-band Synthetic Aperture Radar (SAR) images acquired in the Extra-Wide swath mode; (2) level-1b brightness temperatures measured by Advanced Microwave Scanning Radiometer 2 (AMSR2) onboard the JAXA GCOM-W satellite in all frequencies (6.9, 7.3, 10.7, 18.7, 23.8, 36.5, and 89.0 GHz) and polarizations (vertical and horizontal), and (3) numerical weather prediction parameters from the European Centre for Medium-Range Weather Forecasts (ECMWF) Reanalysis v5 (ERA5) including 2-meter air temperature, skin temperature, total column water vapor, total column cloud liquid water, 10-meter eastward and northward wind components. Additional data including pixel-wise Sentinel-1 incidence angles and distance-to-land maps are also provided for each scene. The maximum temporal offset between Sentinel-1 and AMSR2 acquisitions is 7 hours.

In addition to these input data, the dataset also includes professionally mapped sea ice charts corresponding to each scene, which serve as labeled data for model training. These charts, produced by the Canadian Ice Service (CIS) and the Greenland Ice Service at the Danish

Meteorological Institute (DMI) and following the SIGRID-3 standard, provide a polygon-based map of the amount of SIC, SOD, and FLOE present in each scene. Each chart is aligned temporally and geographically with Sentinel-1 SAR images through projection and rasterization, such that the pixel spacing of the SAR images and the ice charts are the same.

The dataset is available in two versions, the raw version and the ready-to-train (RTT) version. In the RTT version, SAR images are downsampled to 80-meter pixel spacing (from the original 40 meters) using a 2×2 averaging kernel, the land masks are aligned with SAR images, and AMSR2 data is resampled to match the Sentinel-1 geometry. Furthermore, all the features in the scenes are standardized using the mean and standard deviation of the feature (channel). Due to the lower amount of preprocessing required and easier replicability, this is the version used in our experiments.

## 2.2. Sea Ice Targets and Classes

In the RTT version of the AutoICE dataset, the labels for each sea ice parameter (SIC, SOD, FLOE) are extracted from the polygon-based ice charts. First, the dominant ice type for each polygon in the chart is determined based on a 65% partial concentration threshold. Specifically, for a given polygon, the ice type with a partial concentration of at least 65% is assigned as the dominant ice type. Then the corresponding SOD and FLOE of that ice type are assigned to all the pixels within that polygon. This threshold ensures that the identified dominant type accurately represents the primary characteristics of the ice within the polygon. If the partial concentration of none of the ice types within a polygon exceeds the 65% threshold, the polygon is considered ambiguous, and no dominant ice type is assigned. Consequently, the polygon does not contribute specific SOD or FLOE labels to the dataset.

For each polygon, the corresponding SIC, SOD, and FLOE values are assigned to all the pixels within that polygon. SIC values are categorized into 11 classes, ranging from 0% to 100% in increments of 10%. SOD is classified into six classes: open water, new ice, young ice, thin first-year ice (thin FYI), thick first-year ice (thick FYI), and old ice. FLOE is divided into seven classes: open water, cake ice, small floe, medium floe, big floe, vast floe, and bergs.

Table 1 presents a list of the features included in the AutoICE challenge dataset.

Table 1. The list of the features used in our experiments and their resolution in the AutoICE challenge dataset. (*) For Sentinel-1 SAR, the numbers in the table represent pixel spacing not spatial resolution.

| Feature Group | Individual Features | Spatial Resolution | Dataset Resolution |
|---|---|---|---|
| Sentinel-1 SAR | HH | 40 × 40 m (*) | 80 × 80 m (*) |
| | HV (denoised) | | |
| | Incidence angle | | |
| AMSR2 Brightness Temperatures | 6.9 GHz (H, V) | 35 × 62 km | 2 × 2 km |
| | 7.3 GHz (H, V) | 35 × 62 km | |
| | 10.7 GHz (H, V) | 24 × 42 km | |
| | 18.7 GHz (H, V) | 14 × 22 km | |
| | 23.8 GHz (H, V) | 15 × 26 km | |
| | 36.5 GHz (H, V) | 7 × 12 km | |
| | 89 GHz (H, V) | 3 × 5 km | |

3. **Methodology**

*3.1. Input preprocessing*

To ensure maximum replicability, particularly for real-time sea ice segmentation in operational environments, we kept the preprocessing to the minimum amount possible and only applied random cropping to the AutoICE input features. With a similar reasoning, we limited the input feature groups used in our experiments to SAR and AMSR2 groups and did not include the ERA5 or auxiliary features, such as distance to land or location information. In the AutoICE dataset, the Sentinel-1 images and the corresponding AMSR2 scenes have a maximum gap of 7 hours between their acquisition times (Stokholm et al., 2024). Whereas the ERA5 variables have a latency of approximately 5 days, and therefore would not suitable (or available) in a real-time segmentation scenario.

The input features provided in the RTT version of the dataset are of an approximate dimension of 5000×5000 pixels. Due to their size, feeding entire images to a CNN-based model during training could quickly lead to memory issues. Therefore, we extracted 20 random sample patches with a size of 768×768 pixels from each scene to ensure enough coverage for each scene. We selected this patch size generated the best results in a study on the impact of spatial context on sea ice concentration by Stokholm et al. (2022) on a similar dataset. Furthermore, its larger size compared to the 256×256 patch size used by X. Chen et al. (2024 a,b) allows the model to capture a greater spatial context. When generating patches for a scene, if more that 30% of the area of the patch is masked (e.g., contains land), the patch is discarded and another random patch is generated. We used a total of 487 scenes (= 9740 patches) for model training.

We also need to specify the relationship of spatial context to the spatial scale of objects in our study. Spatial context is controlled through the receptive field of the neural network, which we control by varying the atrous rates in the ASPP module. The changing of spatial context may lead to better detection of objects with a larger spatial scale, such as vast floes. We systematically varied the model's receptive field (using different atrous rates in the ASPP module) to understand what level of spatial context was optimal for sea ice objects as observed in each input type (e.g., high resolution SAR vs low resolution PM). Thus, while training patch size ensures enough spatial coverage is presented to the model, the true scale effects in our results come from three related factors: (a) the native resolution and resampling of the input data (Sentinel-1 at tens of meters, AMSR2 at kilometers), (b) the chosen receptive field (atrous rates), which controls whether the model emphasizes finer local details or broader-scale context, (c) the spatial scale of the specific sea ice class.

Since we systematically vary the atrous rates while keeping the patch size constant, our results primarily reflect the effect of spatial context at the feature map level rather than being an artifact of the selected patch size. In other words, the 768-pixel patch itself does not set a fixed scale. Instead, it was a practical choice to preserve enough surrounding area in each training example. To validate our models, we created a hold-out validation set using 24 scenes from the training set. These validation scenes are selected such that the distribution of the classes and the acquisition times are similar to the training set. The validation scenes remained unseen by the models during training, and we used them to determine when to stop training (based on improvement in validation loss). To evaluate our final models, we used the test set provided in the AutoICE challenge which consists of 20 scenes.

We follow the test evaluation strategy of the AutoICE Challenge (Stokholm et al., 2024). where a predefined training and test split is provided. These testing scenes were carefully selected by a committee of sea ice analysts and deep learning experts who organized the aforementioned challenge. These included the Norwegian Computing Center, Danish Meteorological Institute, Technical University of Denmark, Polar View, Nansen Environmental Remote Sensing Center, and European Space Agency. Multiple factors were considered when selecting the test scenes, including seasonal variations, different ice types, and regional variation considerations to ensure a fair and robust test set is provided. Using this test set enables consistency with prior and future work using this benchmark, and allows for a direct comparison of our results.

Figure 1 shows the class distribution of each target in the training and test sets. To calculate the class distributions for the training set, we computed the average of the actual samples (randomly cropped patches) fed to the model across 30 epochs. This excludes the 24 scenes held-out for validation. The invalid column in each category represents the samples that do not meet the 65% threshold for partial concentration. We used the latter condition to discard samples where the number of masked pixels was deemed too high.

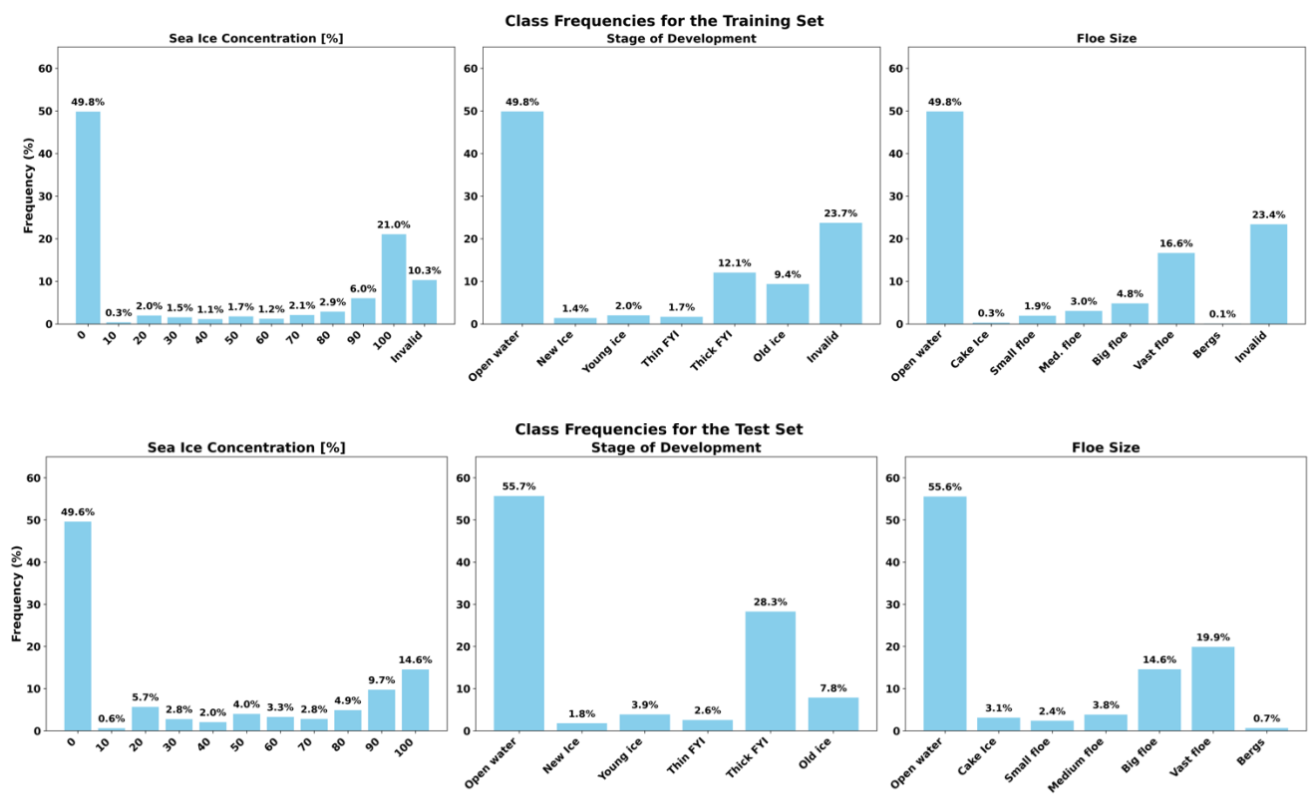

Figure 1. The distribution of individual classes for each task (target). Top: training dataset samples, bottom: test dataset.

*3.2.Model Architecture*

Our proposed model architecture for multi-task sea ice segmentation is based on a modified DeepLab framework (Pires de Lima et al., 2023). We further extended this architecture to segment three distinct properties of sea ice simultaneously (SIC, SOD, and FLOE). Our architecture consists of an encoder and three task-specific decoders, each corresponding to one of the segmentation tasks.

The encoder in this study (unlike Pires de Lima et al. (2023)) is derived from ResNet-101 (He et al., 2016), utilizing only the first three layers of the architecture. This reduces the number of model parameters while maintaining its capacity to extract critical spatial features from the input data. In our initial experiments, in addition to ResNet-101, we tested ResNet-18, ResNet-50 and ResNet-152 and found ResNet-101 to have the best balance between performance and training speed. We used the pretrained version of ResNet-101 (on the ImageNet dataset) for fine-tuning and modified the input to the encoder to handle multiple channels and allow for utilizing different feature groups and channel numbers.

Figure 2 presents the overall schema of our encoder-decoder architecture. In the encoder, the input image is downsampled through a 7×7 convolution and max pooling, followed by three blocks containing 3, 4, and 23 "bottleneck" units respectively. Each bottleneck unit consists of 1×1, 3×3, and 1×1 convolutions each followed by batch normalization and ReLU activation functions. The encoder outputs feature maps from the third layer of ResNet-101, which are subsequently fed into each of the three decoders. The dimensions of these feature maps are 48×48 pixels with 1024 channels.

Each decoder in our model consists of the ASPP module followed by four convolutional layers, each followed by batch normalization, and ReLU activation functions. The ASPP module leverages atrous (dilated) convolutions, which differ from normal convolutions in that they introduce holes between the filter weights by inserting zeros between filter values. The number of zeros inserted between two consecutive filter value is determined by the atrous rate and is equal to r-1 where r is the atrous rate. By adjusting the atrous rate, the filter effectively changes its sampling density—increasing the atrous rate increases the spacing between the sampled points of the input, allowing the filter to capture a larger receptive field. By using atrous convolutions, we can expand the receptive field of the filter without increasing the number of parameters or further downscaling the feature maps (Chen et al., 2018). This allows us to incorporate a wider spatial context while maintaining the spatial resolution of the encoded features. Therefore, the dimension of the decoded feature maps, before the final upsampling layer described below, remains at 48×48 pixel.

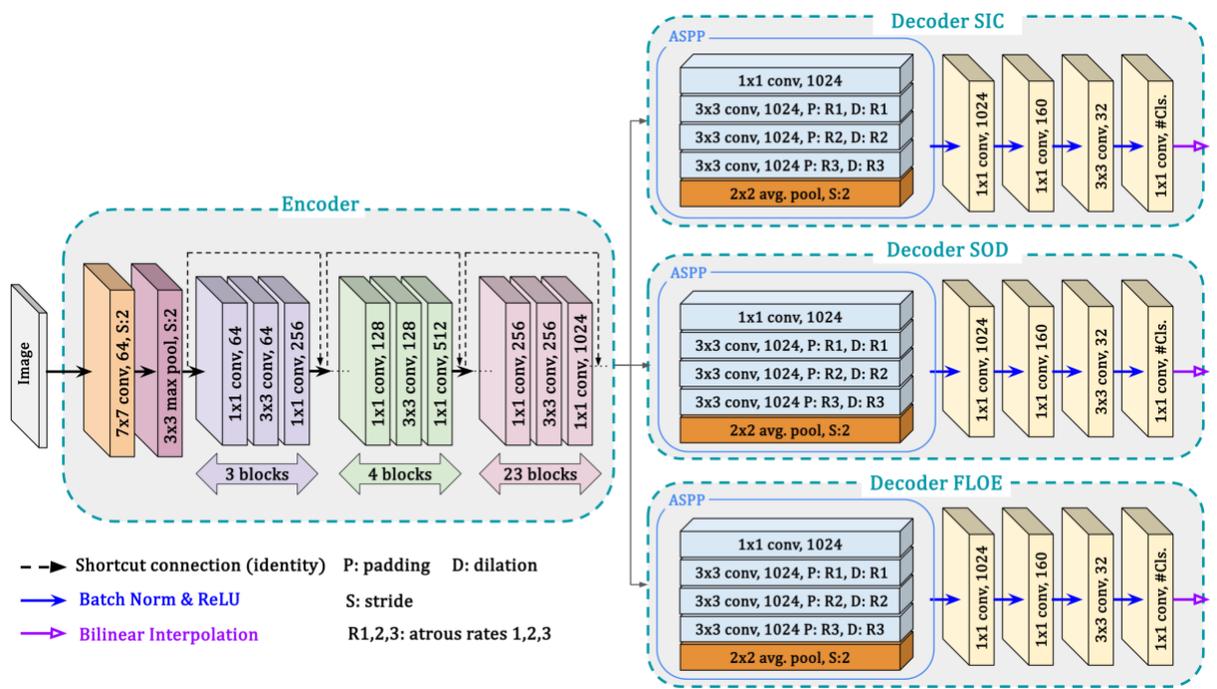

Figure 2. Model architecture. The encoder is the first three blocks (layers) of ResNet-101. The initial 7x7 convolution in the encoder is followed by a 3x3 max pooling layer with a stride of 2. Each block contains a number of bottleneck units that consist of three convolutions. All of the convolutions (represented by conv) in the encoder are followed by batch normalization and the last convolution in each block is followed by ReLU activation after normalization. Each decoder

consists of the ASPP module, four convolutional layers, and an upsampling layer (bilinear interpolation) S represent the stride of convolution. P and D represent the padding and dilatation in the ASPP module, respectively. R1-3 represent atrous rates which also determine P and D.

Each atrous (dilated) convolution layer increase the receptive filed size of its previous layer based on the formula below where for layer $l$, $RF_l$ is the receptive field, $k_l$ is the kernel size (3 in our model), and $D_l$ is the dilation rate, and $RF_{l-1}$ is the receptive field at layer $l$-1 (Figure 3).

$$RF_l = RF_{l-1} + (k_l - 1) \times D_l \tag{1}$$

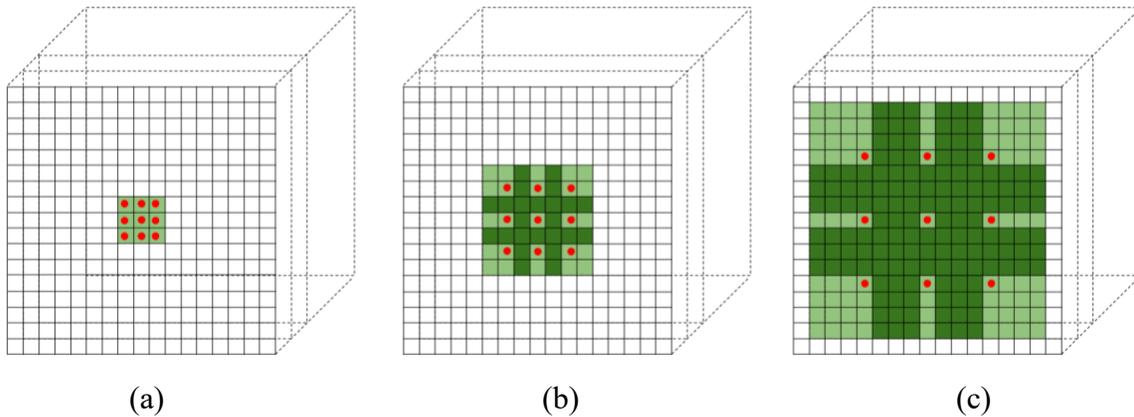

(a)          (b)          (c)

Figure 3. Exponential expansion of CNN's receptive filed based on the dilation rate (originally presented in (Yu and Koltun, 2016). (a) a convolutional kernel of size 3x3 and the receptive field of 3x3 (b) The convolutional kernel with the dilation rate of 2 and receptive field of 7x7 (c) The convolution kernel with the dilation rate of 4 and receptive filed of 15x15. The dashed rectangles in each image represent the different channels of the encoded feature maps (1024 channels in our model).

The ASPP module applies three parallel atrous convolutions with different atrous rates. By doing so, it enables the model to capture information from the encoder-extracted feature maps at multiple spatial contexts and scales simultaneously. We replaced the ASPP module's global average pooling layer with an average pooling layer with 2×2 convolutions and a stride of 2 similar to Pires de Lima et al. (2023). This allows the model, trained on 768×768 patches, to perform inference on full-size SAR images without modification.

During the forward pass, the encoder processes the input pacthes, generating feature maps that are shared among the three decoders. The decoders then process the feature maps at multiple scales denoted by different dilation rates, concatenates the processed multi-scales features, and produce segmentation maps for SIC, SOD, and FLOE, which are then upsampled to the original input resolution using bilinear interpolation. The entire model has a total of 30.45 million trainable parameters. We have made our code publicly available on GitHub (at https://github.com/geohai/segmentation-scale) to ensure the replicability of our results.

This model architecture allows us to regulate the amount of spatial context through the receptive field of convolutional layers, which itself is controlled by varying atrous rates in the ASPP module. The atrous rate serves as a direct mechanism for controlling spatial context within a scene because it dictates the range of spatial relationships that a convolutional filter can capture. Unlike traditional definitions of spatial context in remote sensing that focus on relationships between adjacent pixels in the input domain, our approach captures multi-scale spatial dependencies from the deep feature representations encoded by the model's encoder. Atrous convolutions allow us to expand the receptive field without reducing feature map resolution, enabling the model to incorporate both fine-scale and broader spatial patterns essential for sea ice segmentation. By systematically adjusting the atrous rates, we investigate how spatial context influences the segmentation of different sea ice properties beyond the scale defined by input resolution alone.

### 3.3. Experimental Setup

To examine the effect of spatial context on sea ice segmentation, we designed a set of experiments using different feature groups from Table 1, along with three distinct sets of atrous rates that progressively increase the receptive fields of the CNNs without introducing extra parameters. This design allows us to control the amount of spatial context the model captures. By combining these feature groups with varying atrous rates, we can investigate the impact of spatial context on each feature group individually and explore how these factors interact in distinguishing each sea ice property as well as their combined score.

We selected the following feature groups based on their relevance to sea ice properties and their complementary spatial and spectral characteristics, as demonstrated in previous studies: AMSR2 (18.7, 36.5 GHz in both H &V polarizations), AMSR2 (18.7 V, 36.5 H/V, 23.8 V GHz), AMSR2 (all bands in both H & V polarizations), SAR (HH, HV, Incidence Angle), the fusion of SAR with AMSR2 (18.7, 36.5 GHz in both H &V polarizations) bands, and the fusion of SAR with all AMSR2 bands (in both H &V polarizations). Table 2 provides a list of the above-mentioned groups.

Table 2. The list of feature groups used in our experiments.

| Input Group | Sentinel-1 Features | AMSR2 Features |
| --- | --- | --- |
| 1 | HH, HV, Incidence Angle | - |
| 2 | - | 18.7, 36.5 GHz H/V |
| 3 | - | 18.7 V, 36.5 H/V, 23.8 V GHz |
| 4 | - | 6.9, 7.3, 10.7, 18.7, 23.8, 36.5, 89.0 GHz H/V |
| 5 | HH, HV, Incidence Angle | 18.7, 36.5 GHz H/V |
| 6 | HH, HV, Incidence Angle | 6.9, 7.3, 10.7, 18.7, 23.8, 36.5, 89.0 GHz H/V |

To evaluate the potential benefit of AMSR2 data, we included different combinations of brightness temperatures in our experiments. The AMSR2 (18.7, 36.5 GHz) group balances spatial resolution with reduced sensitivity to atmospheric interference and is traditionally used in sea ice mapping algorithms for both SIC and SOD (Minnett et al., 2019). Moreover, this combination was used by X. Chen et al. (2024 a) whose model achieved the best score in the AutoICE challenge. The AMSR2 (18.7 V, 36.5 H/V, 23.8 V GHz) feature group is used by the Japan Aerospace Exploration Agency (JAXA) Earth Observation Research Center to generate SIC products using the bootstrap method, therefore using the feature group allows us to compare against this method (Comiso and Cho, 2022).

Lastly, using all AMSR2 bands allows for comprehensive coverage of sea ice properties across multiple scales and resolutions, enabling the model to capture both fine details and broader context necessary for segmenting different sea ice properties. This multi-band approach is particularly useful in a multi-task setting, where each sea ice property may require varying levels of spatial and spectral detail. The finer resolution provided by the 89 GHz band is particularly valuable for capturing detailed ice features, despite its susceptibility to atmospheric contamination such as water vapor and cloud liquid water. This band is used by the ARTIST sea ice (ASI) algorithm (Spreen et al., 2008) to provide SIC estimations and in deep-learning-based studies such as (Cooke and Scott, 2019; Radhakrishnan et al., 2021).

While SAR offers much higher resolution, it suffers from signal ambiguity among different types, as well as less frequent revisit times compared to passive microwave. Additionally, the fusion of SAR data with AMSR2 bands allows the model to integrate high-resolution SAR information with broader spatial context provided by the microwave radiometer. This fusion has been shown to improve sea ice segmentation accuracy, particularly for properties like SOD and SIC where both fine details and contextual information are important (Khachatrian et al., 2023; Radhakrishnan et al., 2021; L. Zhao et al., 2023). We chose the specific combination of 18.7 and 36.5 GHz brightness temperatures with SAR data to balance resolution with context. The final feature group combines SAR data with all AMSR2 bands, aiming to maximize the model's ability to capture a wide range of sea ice properties. This feature group allows us to examine how spatial context and multi-frequency data interact in the segmentation of SIC, SOD, and FLOE. Furthermore, we experimented with three distinct sets of atrous rates in the decoder: [6, 12, 18], [12, 24, 36], and [18, 36, 54], referred to as small, medium, and large atrous rates, respectively. The small atrous rates [6, 12, 18] were originally proposed and used by Chen et al. (2017) in the ASPP module of DeepLab V3. The medium atrous rates [12, 24, 36] is the default set of rates used by PyTorch in the implementation of this architecture. Extending these two sets of rates, we used the large rates [18, 36, 54] to allow for a broader spatial context. Together, the three sets (triplets) of atrous rates provide varying receptive fields, allowing the model to capture different levels of spatial context. Smaller atrous rates place more focus on fine details and local context, while larger atrous rates capture more global spatial relationships. This approach enables the model to integrate the diverse spatial patterns observed in sea ice data.

The combination of the above mentioned 6 feature groups and 3 sets of atrous rates results in 18 different experiments. For each experiment, we trained a model using the architecture described in Section 3.2 with the corresponding atrous rates, using the same training specifications which will be described below. We also evaluated all the experiments on the same test set.

### 3.4. Training Details

The models were trained for a maximum of 100 epochs, with an early stopping mechanism applied based on validation loss, with a patience of 10 epochs. We used the Adam optimizer (Kingma and Ba, 2015) with an initial learning rate of 1e-5, which was reduced by a factor of 0.5 after reaching a plateau, with a patience of 4 epochs. Categorical Cross Entropy (CCE) loss was used to optimize all tasks, with the individual losses for the three tasks aggregated to form the total loss during training. The models were trained with a batch size of 8, using a single Nvidia RTX A6000 GPU. We repeated each experiment twice to reduce the randomness introduced during the training process and reported the average of the two runs. Each training and validation epoch took approximately 15 to 16 minutes, and the models reached the stopping criteria after 47 epochs on average.

While statistical significance testing is widely used in inferential statistics, it is not commonly applied in deep learning studies. This is because deep learning models are inherently stochastic, meaning that small variations in initialization, batch sampling, or hyperparameters can lead to slightly different outcomes (Dror et al., 2019). Instead of formal significance tests, deep learning studies typically mitigate randomness by running experiments multiple times and reporting average performance metrics. In our study, we follow this standard practice by conducting each experiment twice and reporting average results. Additionally, our approach is consistent with prior work on sea ice segmentation (X. Chen et al., 2024b, 2024a; Y. Chen et al., 2024). Furthermore, in machine learning, hyperparameter tuning and model selection rarely involve significance testing (Belkin et al., 2019). Instead, models are compared based on validation performance, with the goal of optimizing generalization rather than proving statistical differences between configurations. In our case, the experiments were designed to analyze how receptive field size affects feature learning, following standard deep learning methodologies.

Finally, while some performance differences may appear small, even small but consistent improvements indicate an optimized receptive field configuration for different sea ice properties.

*3.5. Evaluation metrics*

To evaluate the model's performance, we utilized the metrics set in the AutoICE Challenge (Stokholm et al., 2024). The main metric used to evaluate the participants of the challenge is called combined score, which is the weighted sum of three scores, one score per task, and is calculated as:

$$Combined\ Score = \frac{2}{5} \times Score_{SIC} + \frac{2}{5} \times Score_{SOD} + \frac{1}{5} \times Score_{FLOE} \qquad (2)$$

We also used the AutoICE combined score as the main metric to evaluate our experiments. As seen above, the weight of the FLOE score is half the weight of other scores, due to its lower priority for the ice services. The scores for Stage of Development ($Score_{SOD}$) and floe size ($Score_{FLOE}$) are determined using the F1 score and $Score_{SIC}$ is measured using the R² coefficient. For a detailed description of each score please refer to Stokholm et al. (2024).

## 4. Results and Discussion

*4.1. The effect of context on the combined score*

Table 3 and Figure illustrate the impact of the amount of contextual information on the combined score of models trained with different feature groups. The extent of contextual information, controlled by changing the atrous rates in the decoder, determines the receptive field of the CNNs. As shown in Figure , the optimal extent of contextual information required to achieve the best combined score depends on the combination of the features used in the model and the interactions between them. The native resolution (granularity) at which each feature was gathered and the degree of resolution disparity among features used in a model appear to be important factors in determining the optimal atrous rates.

Table 3. The summary of the performance of the different feature groups and atrous rates. All the scores are calculated on the test set.

| Group | Features | Atrous Rates | Test Scores (%) SIC ($R^2$) | SOD (F1) | FLOE (F1) | Combined |
|---|---|---|---|---|---|---|
| 1 | Sentinel-1 SAR | [6, 12, 18] | 75.1 | 74.71 | 70.42 | 74.03 |
| | | [12, 24, 36] | 68.07 | 73.39 | 67.85 | 70.15 |
| | | [18, 36, 54] | 65.97 | 73.04 | 68.42 | 69.28 |
| 2 | AMSR2 (18.7 H/V, 36.5 H/V GHz) | [6, 12, 18] | 72.2 | 75.63 | 68.95 | 72.92 |
| | | [12, 24, 36] | 75.95 | 76.09 | 69.39 | 74.69 |
| | | [18, 36, 54] | 73.99 | 75.61 | 69.1 | 73.66 |
| 3 | AMSR2 (18.7 V, 36.5 H/V, 23.8 V GHz) | [6, 12, 18] | 66.90 | 74.63 | 69.85 | 70.58 |
| | | [12, 24, 36] | 74.58 | 76.39 | 70.07 | 74.40 |
| | | [18, 36, 54] | 72.805 | 75.42 | 70.05 | 73.3 |
| 4 | AMSR2 (All frequencies) | [6, 12, 18] | 75.38 | 74.66 | 69.19 | 73.85 |
| | | [12, 24, 36] | 73.71 | 72.89 | 66.39 | 71.92 |
| | | [18, 36, 54] | 70.60 | 72.94 | 67.24 | 70.87 |
| 5 | Sentinel-1 SAR + AMSR2 (18.7 H/V, 36.5 H/V GHz) | [6, 12, 18] | 86.64 | 75.77 | 73.11 | 79.59 |
| | | [12, 24, 36] | 85.50 | 76.86 | 70.245 | 78.96 |
| | | [18, 36, 54] | 87.05 | 77.15 | 72.42 | 80.17 |
| 6 | Sentinel-1 SAR + AMSR2 (All frequencies) | [6, 12, 18] | **87.34** | **80.66** | **71.885** | **81.58** |
| | | [12, 24, 36] | 84.48 | 79.27 | 71.37 | 79.77 |
| | | [18, 36, 54] | 82.74 | 78.98 | 69.905 | 78.67 |

For the models trained only with SAR features, which have a native resolution of 93×87 meters, decreasing the atrous rates—hence reducing the receptive field—leads to improved performance. The best combined score is achieved by the smallest atrous rates [6, 12, 18], because a smaller receptive field is more capable of capturing local, small-scale information without introducing too much spatial context. Notably, the SAR feature group exhibits the highest variation in the combined score with changing atrous rates, with a 4.07% difference between the best and worst performing models. This indicates the sensitivity of SAR features to the amount of contextual information captured by the model. Therefore, if a model is using only SAR features, the default PyTorch values (the medium rates) may not yield the best results.

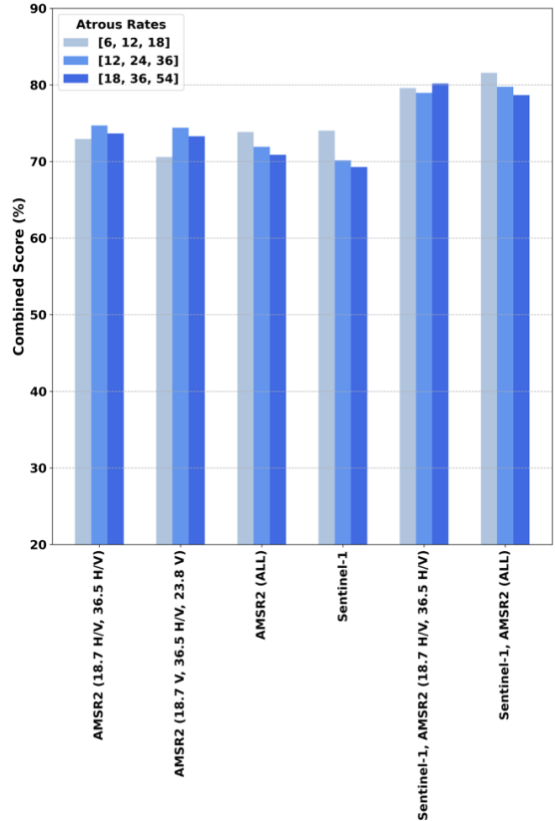

Figure 4. Visual representation of the impact of spatial context on the combined score across different feature groups and atrous rates.

In contrast, AMSR2 features, which have lower (kilometer scale) resolutions, capture varying spatial context across different bands. When SAR features are fused with AMSR2 features, the optimal atrous rates depend on the specific AMSR2 bands included and the variation in their native resolutions. For instance, when all AMSR2 bands are combined with SAR features, the optimal atrous rates are again the smallest rates of [6, 12, 18]. This can be attributed to the complexity introduced by the diverse resolutions of the AMSR2 bands, which range from 35×62 km² for the 6.9 GHz band to 3×5 km² for the 89 GHz band. Given this range of variation, the optimal context seems to be dominated by the finer SAR features. Smaller atrous rates ensure that the model focuses on high-resolution details without being overwhelmed by the coarse context from AMSR2 features.

When all AMSR2 bands are fused with SAR features, the difference between the lowest and highest combined scores is 3.09%. However, when SAR features are fused only with the 18.7 and 36.5 GHz AMSR2 bands, all atrous rates yield similar performance and this difference decreases to 0.89%. The similar ground resolutions of these AMSR2 bands (14×22 km² and 7×12 km², respectively) result in a more consistent response to changes in atrous rates compared to using all AMSR2 bands, which have a wider range of resolutions. As a result, SAR features do not dominate the optimal context as strongly. Even though the differences are small, the best performance is achieved using the largest atrous rates [18, 36, 54], which suggests that larger atrous rates integrate high-resolution SAR information with the broader context provided by the AMSR2 bands more effectively.

Similarly, when only AMSR2 features are used, the optimal atrous rates depend on the interaction between the spatial resolutions of the input bands and the need to balance fine details with broader contextual information. For the combination of 18.7 GHz and 36.5 GHz bands, medium atrous rates [12, 24, 36] yield the highest combined score. When the 23.8 GHz band is added to this combination, medium atrous rates again produce the best results. These atrous rates effectively integrate the different levels of context and enhance the model's capability to capture the multi-scale features essential for segmentation tasks.

However, when all AMSR2 bands are utilized, including the high-resolution 89.0 GHz band (3 km × 5 km), the smallest atrous rates [6, 12, 18] yield the highest performance, while medium and large atrous rates result in similar combined scores. This suggests that the higher resolution data from the 89.0 GHz band becomes dominant, requiring the model to focus on fine-grained details without being overwhelmed by broad-scale context. In this case, larger atrous rates likely introduce too much contextual information, overwhelming the fine context captured by the higher resolution bands and leading to a decline in performance.

As illustrated in Figure , an important observation is that the fusion of SAR and AMSR2 features consistently yields higher combined scores compared to using a single source of input, regardless of the atrous rates applied. However, as noted before, AMSR2 offers more frequent revisit times than S-1.

*4.2. The effect of context on individual sea ice properties*

As evident in Table 3, the sensitivity of individual segmentation tasks (SIC, SOD, FLOE) to spatial context varies. Based on our experiments, the best performing atrous rates depend on the dominant feature(s) in the input data and their influence on each sea ice property. Figure presents the impact of atrous rates on the individual tasks, which are discussed separately below.

- **Sea Ice Concentration (SIC)**

The SIC R² scores, as shown in Figure A, generally follow trends similar to the combined score across most feature groups, with the exception of the AMSR2 all-bands group. In this case, medium atrous rates yield better results compared to large rates, whereas large atrous rates slightly outperform medium rates in the combined score metric. This similarity between SIC and combined score trends may be due to the calculation method of the combined score and the weight assigned to the SIC scores. The fact that SIC $R^2$ scores are generally higher than SOD and FLOE means that the combined score is influenced more by SIC than the other two tasks.

For the AMSR2 (18.7, 36.5 GHz bands) and AMSR2 (18.7 V, 36.5 H/V, 23.8 V GHz bands) groups, medium atrous rates [12, 24, 36] yield the highest SIC performances, indicating a balanced use of spatial context. SAR-only features show the highest SIC performance with the small atrous rates, while the fusion of SAR with AMSR2 (18.7, 36.5 GHz bands) performs best with larger atrous rates, suggesting that broader contextual information benefits this combination. This consistent trend across SIC and combined scores underscores the balance required between spatial resolution and contextual information for optimal sea ice segmentation. Additionally, similar to the combined scores, the fusion of SAR and AMSR2 features consistently yields higher SIC performances than using a single input source, regardless of the atrous rates applied.

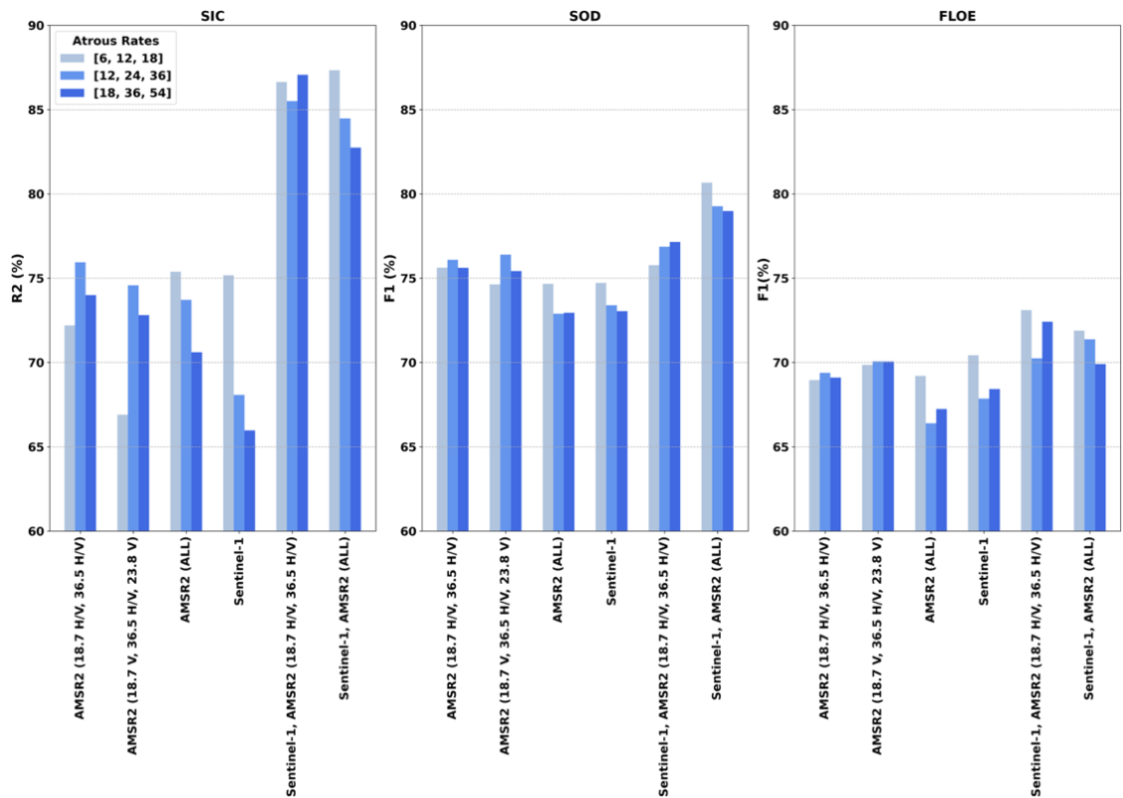

Figure 5. Visual representation of the impact of spatial context on each task across different feature groups and atrous rates. A) Sea Ice Concentration (SIC), B) Stage of Development (SOD), C) floe size (FLOE).

- **Stage of Development (SOD)**

The impact of the atrous rates on SOD F-1 scores presents different patterns compared to SIC and combined scores. The medium atrous rates yield the best performance in the SOD segmentation task for. This indicates that moderate spatial context provided by a medium receptive field is optimal for differentiating development stages, whereas small or large receptive fields might focus too narrowly on fine details or too broadly on large-scale features, respectively.

Furthermore, the fusion of SAR with all AMSR2 bands achieves the highest SOD scores across all atrous rates, unlike the performance observed for SIC and combined scores. This discrepancy suggests that SOD segmentation benefits from the broader spatial context provided by diverse AMSR2 data. Additionally, the model trained with only AMSR2 (18.7 & 36.5 GHz) bands achieves the second highest SOD score across all atrous rates, outperforming the model trained with SAR features consistently. This suggests that, when leveraged in modern deep learning frameworks, these bands are beneficial in SOD mapping in addition SIC mapping which is the most common application of these bands in the literature (despite their lower resolution).

- **Floe Size (FLOE)**

FLOE segmentation scores show that the smallest atrous rates [6, 12, 18] yield the best performances across most feature groups, with exceptions in AMSR2 (18.7 & 36.5 GHz) and AMSR2 (18.7 V, 36.5 H/V, 23.8 V GHz) groups. This suggests that smaller receptive fields are more effective in capturing floe size variations, while larger contextual information provided by medium and large atrous rates may not be as crucial for this task. This finding contrasts with SOD segmentation, where broader spatial context has a beneficial impact.

For AMSR2 (18.7 & 36.5 GHz) bands, medium atrous rates produce the highest FLOE score, while small and large rates perform similarly but slightly lower. This trend aligns with combined scores, SIC, and SOD where medium rates yield the best results. A similar trend is observed with AMSR2 (18.7 V, 36.5 H/V, 23.8 V GHz) bands, where medium and large atrous rates perform comparably, with medium slightly better. This aligns with the combined scores, SIC, and SOD where medium atrous rates outperform small and large rates. For all AMSR2 bands, both small and large atrous rates yield better performance compared to medium rates by a notable difference. In addition, small atrous rates yield the best performance as was the case for SIC and combined scores.

Unlike SIC and SOD, different feature groups achieve comparable (but lower across the board) performances in FLOE segmentation, with smaller differences in F-1 scores across feature groups. However, similar to SIC and the combined scores, the fusion of SAR and AMSR2 features (including all bands and 18.7 & 36.5 GHz) consistently achieves the highest scores across all atrous rates.

### 4.3. Grad-CAM analysis

Grad-CAM (Gradient-weighted Class Activation Mapping) is a visualization and interpretability technique used to understand which parts of an input image contribute most to a deep learning model's decision, particularly for CNNs (Selvaraju et al., 2017). It works by computing the gradient of the target class score with respect to the feature maps of a convolutional layer. These gradients are then averaged per feature map to produce weights for the importance of each feature map, which are used to create a weighted combination of the feature maps. This combination generates an activation map (heatmap) that highlights the regions in the image that had the most influence on the model's decision.

Analyzing Grad-CAM heatmaps is helpful in understanding how different atrous rates affect the model's attention and receptive field and can help with the interpretability for specific classes. We selected the model trained with SAR features for this analysis, as it demonstrates notable performance variation with changes in the receptive field (atrous rates), and that SAR is the primary source for recent research in automated sea ice mapping. We then generated Grad-CAM heatmaps for one of the test scenes (20210506T075557_dmi) using the gradients of the last convolutional layer of each decoder.

Figure presents the ground truth labels provided for the selected scene, alongside the model predictions generated by the models trained with SAR features with different atrous rates. This scene contains a significant amount of old ice and a small amount of thick FYI. It also includes ice with varying concentrations (30%, 70%, 80%, 90%, and 100%), with the 90% concentration ice being the most prevalent class. Additionally, it contains small, medium, big, and vast floe ice. Considering the close FLOE scores for this scene (0.82% variation across different atrous rates), we focus on SOD and SIC for this analysis, specifically on the most prevalent class in each task for the scene, i.e., old ice for SOD and 90% concentration for SIC.

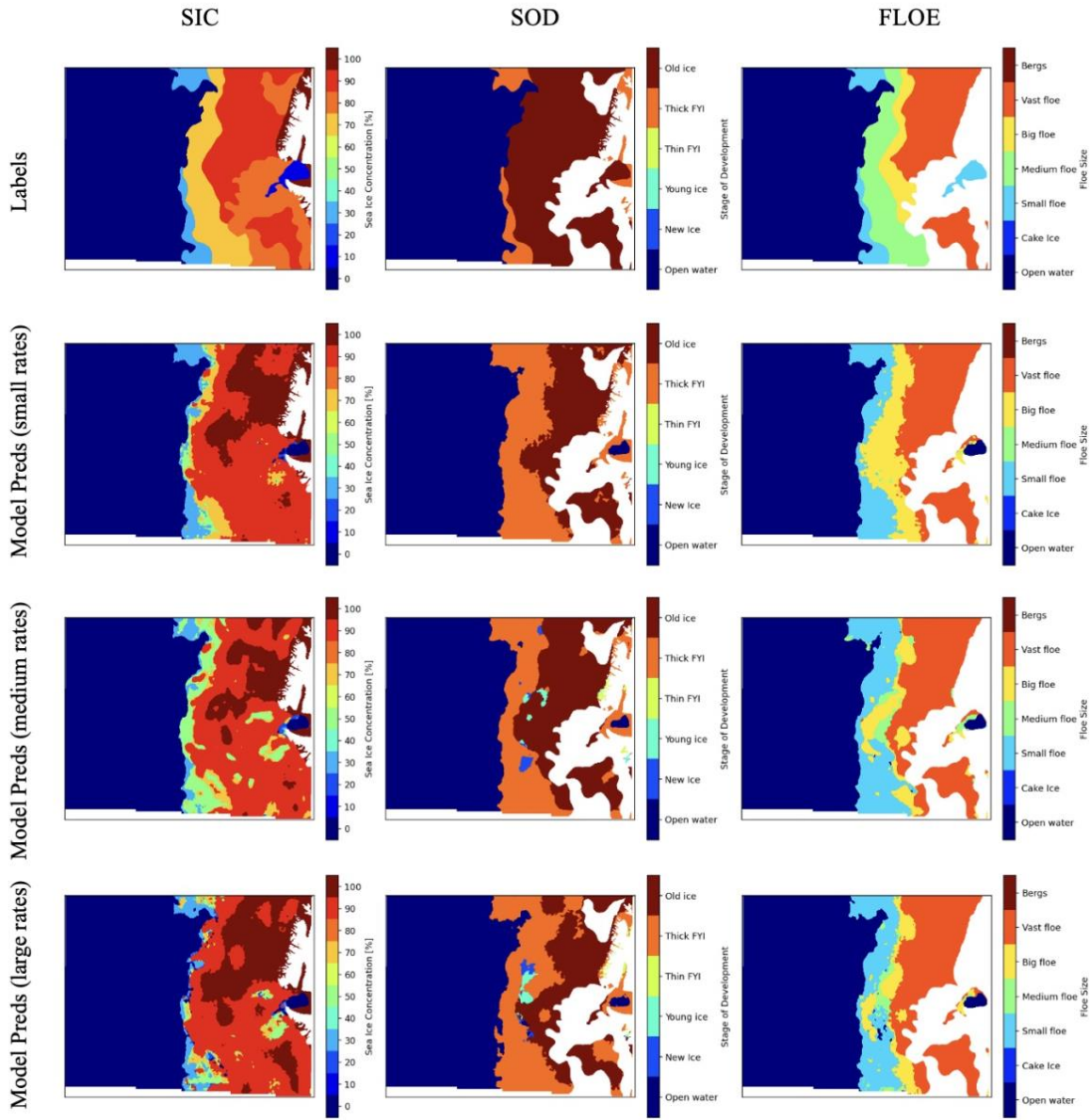

Figure 6. *Manually* labelled sea ice charts and model predictions for scene 20210506T075557_dmi. Left column: Sea Ice Concentration (SIC), middle column: Stage of Development (SOD), right column: Floe size (FLOE). From top to bottom: A) Ground truth labels. B) Model predictions generated by the model trained with SAR features and [6,12,18] atrous rates. C) model predictions generated by the model trained with the same features and [12,24,36] atrous rates. D) model predictions generated by the model trained with the same features and [18,36,54] atrous rates. The white regions in the SOD and FLOE columns are masked (i.e. contain land).

Figure illustrates the Grad-CAM heatmaps for the old ice SOD class. The heatmap with the smallest atrous rates [6, 12, 18] (top row), has a more diffused (less crisp) pattern compared to other rates, which might be due to the fact that the model focuses on finer, localized ice features as opposed to broader ice regions. This map shows more regions with medium activation values (green-colored pixels) compared to other heatmaps, and the model misclassifies old ice towards the bottom-center of the scene as thick FYI. The activation values near the Marginal Ice Zone (MIZ) are higher compared to the other two maps, especially in the bottom half of the image. The heatmap for medium atrous rates [12, 24, 36] (middle row) shows a less diffused activation pattern and contains more pixels with maximum activation values compared to small and large rates, indicating higher model confidence in its classification. Considering that the medium atrous rates capture a mix of fine details and larger contextual information, this balanced approach likely explains why these rates often yield optimal performance for SOD segmentation. Even though there are sporadic regions of old ice misclassified as young ice, this model has performed better in correctly classifying old ice towards the bottom- and top-center of the scene. The activation values at the ice edge also remain high, almost like the small atrous rates.

In the heatmap with the largest atrous rates [18, 36, 54] (bottom row), the activation pattern becomes stronger locally. This suggests that the model is incorporating more contextual information at the cost of some finer details which has led to a reduction in the identification of old ice regions. More old ice pixels are misclassified as young and new ice towards the center, and notably more old ice pixels are misclassified as thick FYI towards the bottom- and top-center of the scene compared to the medium atrous rates. The activation values at the ice edge are lower, meaning there is potential for misclassification in those regions. As the atrous rates increase, the effect of banding thermal noise becomes less noticeable (evident as a vertical disturbance line towards the left side of the overlaid images), while the effect of wind roughness (also towards the left side of the overlaid images) appears more prominent.

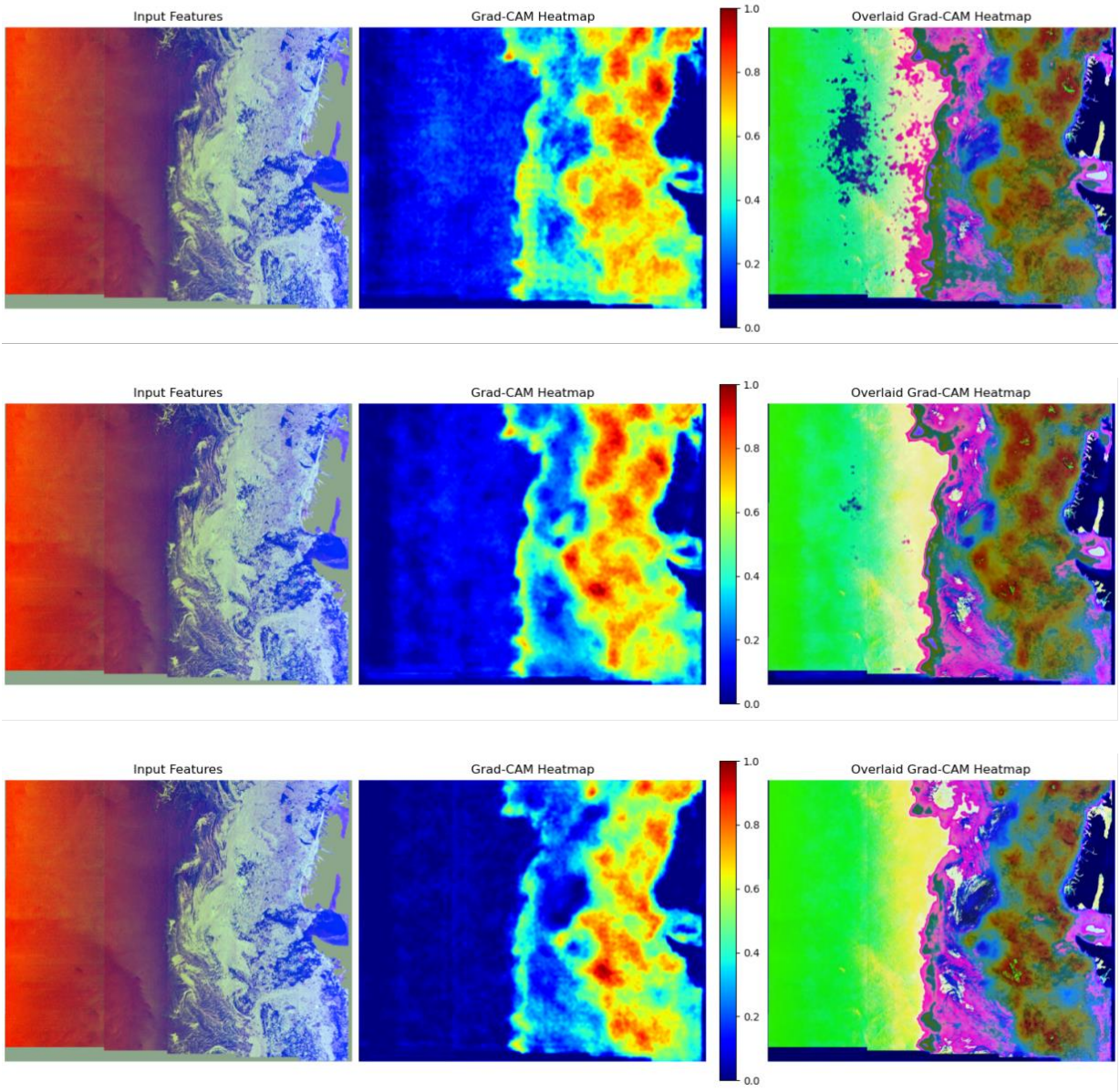

Figure 7. Grad-CAM Heatmaps from the last convolutional layer of the SOD decoder for Stage of Development **(SOD) class Old Ice** for scene 20210506T075557_dmi. Left column: input features (SAR, i.e., HH, HV, Incidence Angle). Middle column: Grad-CAM heatmap generated for each model. Right column: input features overlaid by the respective heatmap. From top to bottom: models trained with [6,12,18], [12,24,36], and [18,36,54] atrous rates respectively.

Figure illustrates the Grad-CAM heatmaps for the 90% concentration SIC class. Compared to the old ice class for the SoD task, all heatmaps for this class in the SIC task have higher average activation values, indicating that the model is more confident in classifying this class (whether correctly or incorrectly). The model also achieved higher average SIC scores on the test scene compared to SOD scores. Moreover, the activation values around the MIZ are higher compared to the old ice class and as seen in Figure , and the ice-water boundary classification is more accurate for 90% concentration class compared to the old ice class. Put differently, a model trained on classifying SIC may perform better than a model performed on classifying SOD if the detection of ice edge is of importance.

Similar to the old ice class, the heatmap with the smallest atrous rates (top row) focuses on fine, localized details of the high concentration regions. This heatmap has more pixels with high activation values (red-colored pixels) compared to the other two atrous rates and has generated a more uniform region of pixels classified as 90% concentration. Although some pixels with 70%, 80%, and 100% concentration are misclassified as 90%, this model achieves the best overall SIC $R^2$.

The heatmap for medium atrous rates (middle row) has more pixels with the maximum activation value (dark-red pixels), particularly near the ice edge and has misclassified these pixels with mostly 70% concentration as 90%. In the heatmap with the largest atrous rates (bottom row), there are fewer pixels with high activation values, indicating a lower confidence in classification. In particular, the region with low activation value towards the upper-right corner of the image is misclassified as 100% ice. Though this misclassification is present in other heatmaps, the affected area is larger in the heatmap with the largest atrous rates due to a lower average activation value. This behavior mirrors that of the old ice class, where larger atrous rates captured more context at the expense of detail. However, the SIC class shows slightly better retention of detail at larger atrous rates compared to the old ice class.

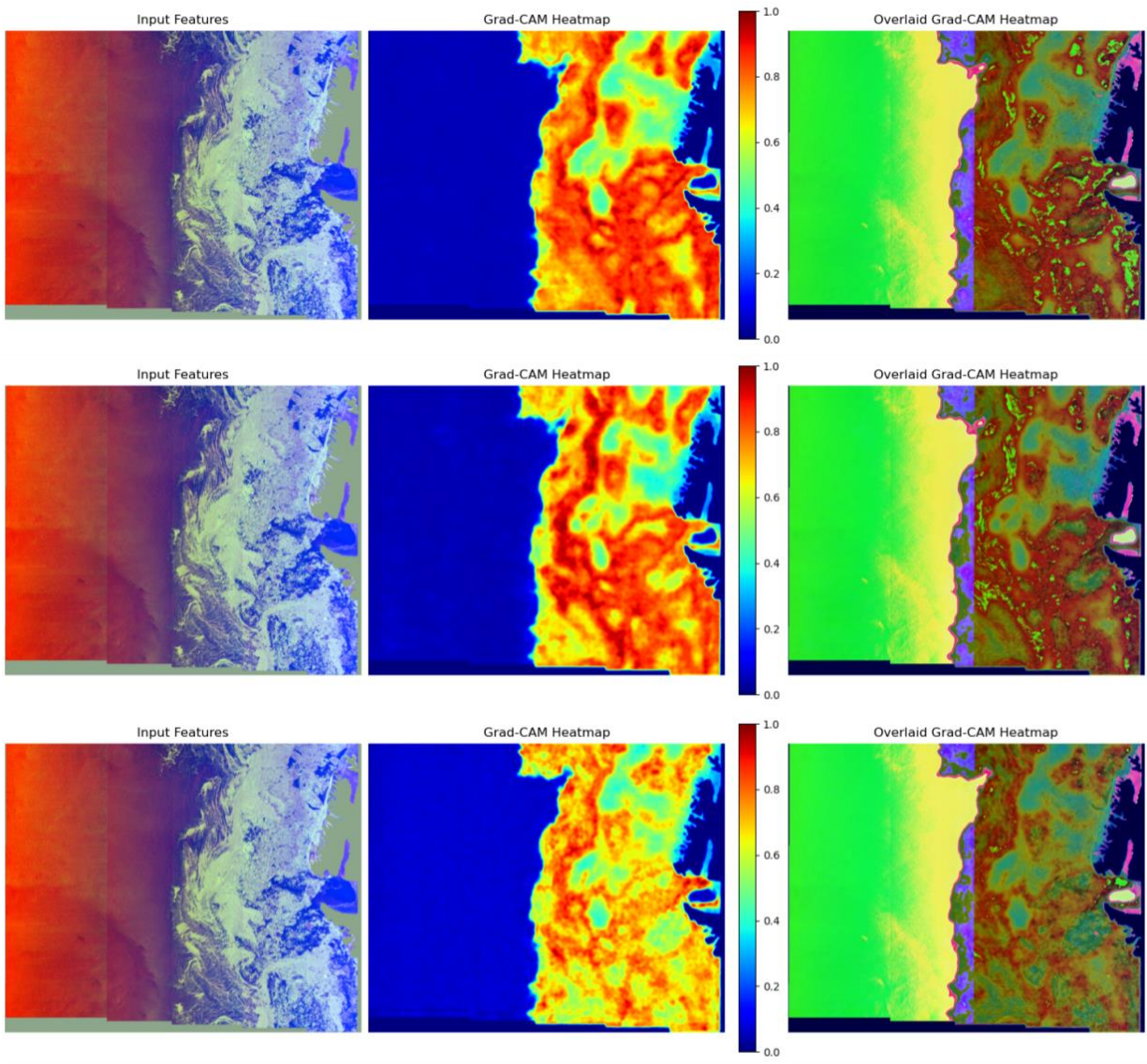

Figure 8. Grad-CAM heatmaps from the last convolutional layer of the SIC decoder for the sea ice concentration **(SIC) class 90%** for scene 20210506T075557_dmi. Left column: input features for all rows (SAR, i.e., HH, HV, Incidence Angle). Middle column: Grad-CAM heatmap generated for each model. Right column: input features overlaid by the respective heatmap. From top to bottom: models trained with [6,12,18], [12,24,36], and [18,36,54] atrous rates respectively.

Overall, both classes exhibit similar trends with respect to atrous rates when only high-resolution SAR is used as input: smaller rates focus on fine details, medium rates balance detail and context, and larger rates capture broad context and likely to miss smaller features. The primary difference lies in the degree of detail retention at larger atrous rates, with the SIC class showing a slight advantage over the SOD class in maintaining finer details within the broader contextual information. This visual comparison shows the impact of atrous rates on different sea ice properties and individual classes, highlighting the importance of selecting appropriate atrous rates based on the specific segmentation task.

### *4.4. Discussion and Comparative Analysis*

In this study, we explored the effect of atrous rates on individual sea ice properties and classes within a multi-task, multi-class segmentation setting. We also investigated the performance of different feature groups, with features from different sensors and channels with different resolutions. In addition to the atrous rates presented here, we tested [3, 6, 9] and [24, 48, 72] atrous rates. The models trained with the smallest atrous rates demonstrated comparable performance to those using [6, 12, 18], but did not provide further insights than those presented above. The substantially larger [24, 48, 72] atrous rates caused models to be unstable, occasionally failing to converge. Even in cases where convergence occurred, the results did not perform as well as the small, medium and large atrous rates presented here, with a combined score of only 57.42%. This indicates that overly large receptive fields may lead to too large a context, ultimately hurting model performance.

Additionally, while this study employed uniform atrous rates across all targets (SIC, SOD, FLOE), our findings suggest that each target responds differently to the amount of context provided by varying atrous rates. Experimenting with different atrous rates for each target could yield more optimized results; however, a systematic conclusion would require extensive experimentation and significant computational resources, which were beyond the scope of this study. Nonetheless, our findings can guide further investigation into target-specific atrous rate configurations.

As shown in Table 3, the performance differences across atrous rates are minimal when the native resolution of the input features is lower. For example, when all AMSR2 bands are used as input features, the combined scores remain close and comparable across all atrous rates. This outcome is expected, as the optimal spatial context for a model is related to the context inherently captured by the feature set. However, when higher-resolution SAR images are used as the only input or fused with lower resolution passive microwave, the selection of atrous rates has noticeable effect on performance.

Compared to a similar study on the AutoICE challenge dataset in X. Chen et al. (2024 b), our modified DeepLabV3 architecture (in conjunction with various atrous rates) achieves a better performance when similar input features are used. Specifically, our model trained with Sentinel-1 SAR features fused with all AMSR2 brightness temperatures achieves a combined score of 82.18%, SIC $R^2$ of 88.4%, SOD F-1 of 80.78%, and FLOE F-1 of 72.53% (with the small atrous rates). The U-Net-based model trained with similar features in (X. Chen et al., 2024 b), albeit excluding Sentinel-1 incidence angle feature, achieved a combined score of 81.15% (SIC $R^2$ of 87.28%, SOD F-1 of 78.43%, and FLOE F-1 of 74.33%). They used Monte Carlo cross-validation where the model was trained and validated 30 times (with different random selections of validation scenes) and reported the average results across these iterations. They also downsampled the input images by a factor of 10.

We also compare our results against the MFDA model proposed by (Y. Chen et al., 2024) that has performed spatial cross-validation on the AutoICE challenge dataset. The capability of a model to generalize across different regions is important in operationalizing automated sea ice mapping. Geographic generalizability has been a common issue in remote sensing tasks, specifically sea ice mapping, as trained models tend not to generalize when applied to different geographic regions. This can lead to reduced reliability in real-world use. Y. Chen et al. (2024) trained their model on the Canadian subset of the AutoICE benchmark, which has 197 scenes as the source domain, and tested on the Greenland subset of 315 scenes as the target domain. We follow the same train/test splits and use the same group of features, with atrous rates of [6, 12, 18] for the decoder head.

Our model achieved higher performance in all three tasks (92.86% vs 78.0% $R^2$ for SIC; 85.38% vs 75.8% SOD F-1 score 88.68% vs 79.8% FLOE F-1 score) as well as the combined score (89.03% vs 77.48%) on the Greenland test subset, even though (Y. Chen et al., 2024) used a larger pre-training dataset. This contrast underscores the trade-off between broad cross-scene adaptability, as pursued by MFDA, and the fine-tuned multi-scale context design of our model. Given that our objective was not to achieve the highest overall score, but rather to isolate and explore the impact of context and scale on multi-task sea ice segmentation, we avoided using ensembling or data pre-processing techniques used in X. Chen et al. (2024 b). Furthermore, by enabling experimentation with different atrous rates, our proposed architecture provides a valuable foundation for investigating the impact of contextual information on segmentation performance.

The findings of Stokholm et al. (2022), using a different training dataset but with similar characteristics, suggest that expanding the receptive field of CNNs improves model performance for classifying SIC from SAR images (without using AMSR2 features). They increased the size of the receptive fields by adding more levels to both the encoder and decoder of a U-Net architecture. In our experiments using SAR bands as input, expanding the receptive field by using larger atrous rates led to a decrease in all tasks. This difference is likely due to the different approaches used to increase the receptive field. Therefore, any direct comparison is challenging. Further research is necessary to fully understand these differences, ideally by applying both methods to the same dataset.

While we have systematically explored the effect of context on sea ice segmentation by changing the receptive field, this research primarily investigates how varying receptive fields (spatial context) affect segmentation performance. Although scale is inherently related to the size of these receptive fields, the study does not directly focus on analyzing how segmentation performance varies with different input resolutions or across distinct geographic scales. Therefore, while multi-scale features are captured by varying atrous rates, the research does not fully address the broader concept of scale as it relates to spatial resolution or geographic scope.

5. **Conclusion**

In this study, we systematically investigated the impact of spatial context on the segmentation of different sea ice properties using a multi-task model. By varying the atrous rates within the Atrous Spatial Pyramid Pooling (ASPP) module, we were able to control the receptive fields of the model and assess how different scales of spatial context affect segmentation performance across these sea ice properties.

Our experiments show that the optimal amount of spatial context, as determined by the atrous rates, is highly dependent on the feature groups used and the specific sea ice property being segmented. For high-resolution Sentinal-1 SAR data, smaller atrous rates were most effective, capturing fine-grained details crucial for accurate segmentation. If only lower resolution AMSR2 features are used, the best atrous rates depend on the specific frequencies incorporated, and therefore, no conclusive rate can be recommended for all bands. However, if both S1 SAR and AMSR2 features are fused in one model, still, small atrous rates result in best overall performance. Larger atrous rates often introduced excessive spatial context, leading to diminished performance by overwhelming high-resolution information with coarse-scale features. This is not a complete surprise, given that SAR sensors provide the most useful information for sea ice mapping, compared to the much lower resolution AMSR2 data, and for SAR, smaller atrous rates yield the best results. Still, the performance of deep learning models leveraging only AMSR2 is impressive and indicates an opportunity to replace existing algorithms such as Bootstrap (Comiso and Cho, 2022) with deep learning methods.

These findings underscore the importance of tailoring the receptive field size to the specific characteristics of the input data and the sea ice properties of interest. While our study employed uniform atrous rates across all targets, the results suggest potential benefits from using target-specific atrous rates to further optimize segmentation performance. Future work could explore this direction, along with the integration of additional features or more sophisticated data fusion techniques, to enhance the robustness and accuracy of automated sea ice mapping.

In this work, all input features are resampled to a common grid before being fed into the model. Our comparison of various future combinations, including several variations of early-, mid-, and deep-fusion strategies (CITATION MASKED- SAME AUTHORS) shows that this approach leads to the best results. This approach has been applied in deep learning-based sea ice segmentation and classification studies (e.g., Stokholm et al., 2024; X. Chen et al., 2024a; Khachatrian et al., 2023), as it allows models to leverage both high-resolution SAR backscatter and lower-resolution passive microwave brightness temperatures without requiring complex deep fusion mechanisms. Future work could explore the effect of alternative fusion strategies on capturing spatial context and scale, such as deep feature fusion or resolution-adaptive architectures, to further optimize multi-source sea ice segmentation.

The findings of this study underscore the importance of multi-scale spatial context in remote sensing applications, providing insights applicable to a range of geospatial machine learning tasks. By tailoring receptive fields to match the distinct characteristics of input data and segmentation targets, this approach offers a path forward in handling the spatial variability encountered in large-scale Earth observation data. In fields such as urban planning, agriculture, and environmental monitoring, where data from multiple sensors are frequently integrated, this study's insights highlight the potential for enhanced model performance through optimized receptive fields and feature fusion. Future research might explore adaptive or task-specific receptive field settings to extend these benefits, and thus, enhancing the capacity of deep learning models to capture complex spatial patterns across diverse applications.

2019, Association for Computational Linguistics, Florence, Italy, pp. 2773–2785. https://doi.org/10.18653/v1/P19-1266

He, K., Zhang, X., Ren, S., Sun, J., 2016. Deep Residual Learning for Image Recognition, in: *Proceedings of the IEEE Conference on Computer Vision and Pattern Recognition*. pp. 770–778.

Huang, Y., Ren, Y., Li, X., 2024. Deep learning techniques for enhanced sea-ice types classification in the Beaufort Sea via SAR imagery. *Remote Sensing of Environment* 308, 114204. https://doi.org/10.1016/j.rse.2024.114204

Ivanova, N., Pedersen, L.T., Tonboe, R.T., Kern, S., Heygster, G., Lavergne, T., Sørensen, A., Saldo, R., Dybkjær, G., Brucker, L., Shokr, M., 2015. Inter-comparison and evaluation of sea ice algorithms: towards further identification of challenges and optimal approach using passive microwave observations. *The Cryosphere* 9, 1797–1817. https://doi.org/10.5194/tc-9-1797-2015

Khachatrian, E., Dierking, W., Chlaily, S., Eltoft, T., Dinessen, F., Hughes, N., Marinoni, A., 2023. SAR and Passive Microwave Fusion Scheme: A Test Case on Sentinel-1/AMSR-2 for Sea Ice Classification. *Geophysical Research Letters* 50, e2022GL102083. https://doi.org/10.1029/2022GL102083

Kingma, D.P., Ba, J., 2015. Adam: A Method for Stochastic Optimization, in: Bengio, Y., LeCun, Y. (Eds.), *3rd International Conference on Learning Representations, ICLR 2015, San Diego, CA, USA, May 7-9, 2015, Conference Track Proceedings*.

Li, M., Zang, S., Zhang, B., Li, S., Wu, C., 2014. A Review of Remote Sensing Image Classification Techniques: the Role of Spatio-contextual Information. *European Journal of Remote Sensing* 47, 389–411. https://doi.org/10.5721/EuJRS20144723

Ma, T., Chen, X., Xu, L., Ma, P., Yu, P., 2025. MFGC-Net: Bridging and fusing multiscale features and global contexts for multi-task sea ice fine segmentation. *IEEE Journal of Selected Topics in Applied Earth Observations and Remote Sensing* 1–21. https://doi.org/10.1109/JSTARS.2025.3551976

Malmgren-Hansen, D., Pedersen, L.T., Nielsen, A.A., Kreiner, M.B., Saldo, R., Skriver, H., Lavelle, J., Buus-Hinkler, J., Krane, K.H., 2021. A Convolutional Neural Network Architecture for Sentinel-1 and AMSR2 Data Fusion. *IEEE Transactions on Geoscience and Remote Sensing* 59, 1890–1902. https://doi.org/10.1109/TGRS.2020.3004539

Minnett, P.J., Alvera-Azcárate, A., Chin, T.M., Corlett, G.K., Gentemann, C.L., Karagali, I., Li, X., Marsouin, A., Marullo, S., Maturi, E., Santoleri, R., Saux Picart, S., Steele, M., Vazquez-Cuervo, J., 2019. Half a century of satellite remote sensing of sea-surface temperature. *Remote Sensing of Environment* 233, 111366. https://doi.org/10.1016/j.rse.2019.111366

Nagi, A.S., Kumar, D., Sola, D., Scott, K.A., 2021. RUF: Effective Sea Ice Floe Segmentation Using End-to-End RES-UNET-CRF with Dual Loss. *Remote Sensing* 13, 2460. https://doi.org/10.3390/rs13132460

Pires de Lima, R., Vahedi, B., Hughes, N., Barrett, A.P., Meier, W., Karimzadeh, M., 2023. Enhancing sea ice segmentation in Sentinel-1 images with atrous convolutions. *International Journal of Remote Sensing* 44, 5344–5374. https://doi.org/10.1080/01431161.2023.2248560

Radhakrishnan, K., Scott, K.A., Clausi, D.A., 2021. Sea Ice Concentration Estimation: Using Passive Microwave and SAR Data With a U-Net and Curriculum Learning. *IEEE Journal*
45